\documentclass[letterpaper, 10 pt, conference]{ieeeconf}
\usepackage{xcolor}
\usepackage{graphicx}
\usepackage{adjustbox}
\usepackage{booktabs}
\usepackage{bm}
\usepackage{amsfonts}
\usepackage{float}
\usepackage{newtxtext}
\usepackage{amsmath,amssymb,bm}

\newcommand{\SHshift}{\gamma_{\mathrm{gain}}}
\newcommand{\ashift}{\gamma_{\mathrm{shift}}}

\IEEEoverridecommandlockouts                              

\title{\LARGE \bf
Directionally Factorized Illumination with Spherical Harmonic Rotation
}
\author{Yimeng Dou$^{1}$, Adam M. Saunders$^{1}$, Andre T. S. Hucke$^{1}$, Michael E. Kim$^{2}$, Adam W. Anderson$^{3-5}$, \\ Eric R. Tkaczyk$^{1,3,6,7}$, Bennett A. Landman$^{1-5}$, Daniel Moyer$^{2}$
\thanks{*This work was supported by the National Institutes of Health under award numbers U54DK134302, U54EY032442, R01EB017230, R01HL169944. This work was also partially supported by VA MERIT award I01 CX002721 from the United States Department of Veterans Affairs Clinical Science R\&D Service, and by the Alzheimer’s Disease Sequencing Project Phenotype Harmonization Consortium (ADSP-PHC) that is funded by NIA (U24 AG074855, U01 AG068057 and R01 AG059716). } 
\thanks{$^{1}$ Department of Electrical and Computer Engineering, Vanderbilt University, Nashville, TN, USA; $^{2}$ Department of Computer Science, Vanderbilt University, Nashville, TN, USA; $^{3}$ Department of Biomedical Engineering, Vanderbilt University, Nashville, TN, USA; $^{4}$ Department of Radiology and Radiological Sciences, Vanderbilt University Medical Center, Nashville, TN, USA; $^{5}$ Vanderbilt University Institute of Imaging Science, Vanderbilt University Medical Center, Nashville, TN, USA; $^{6}$ Department of Dermatology, Vanderbilt University Medical Center; $^{7}$ Veterans Affairs Tennessee Valley Healthcare System}}

\begin{document}
\maketitle
\thispagestyle{empty}
\pagestyle{empty}

\begin{abstract}
Many real-world capture scenarios involve an active light source rigidly attached to the camera: camera flashes for dark-environment imaging, underwater vehicle lights, and endoscopic/laparoscopic medical procedures. As the camera moves, the co-mounted light moves with it, causing the illumination direction at each surface point to change across views. Neural rendering methods (NeRF, Gaussian Splatting) encode this pose-dependent illumination as surface coloration, producing appearance inconsistencies and making it harder to recover the underlying scene. We present \textbf{DaFISH}, which can be plugged into any novel-view synthesis (NVS) pipeline such as NeRF. It factorizes per-point emitted color into surface albedo and a spherical harmonic (SH)-represented illumination factor. During training, the SH illumination is evaluated directly, enabling stable joint optimization of albedo, lighting, and geometry. \textbf{DaFISH} provides analytical generalization of illumination to arbitrary camera poses without retraining. Building on the closed-form SH-to-irradiance conversion, the training pipeline is fully differentiable from SH evaluation through volume rendering. 
Averaged over the held-out frames of the five scenes, \textbf{DaFISH} attains the best mean PSNR (38.91\,dB), SSIM (0.9613), and LPIPS (0.0582) among all compared methods. It improves over the strongest splatting baseline as well as the closest NeRF-based baseline, while recovering the scene consistently across views compared to prior methods.
\end{abstract}

\begin{figure}[t]
    \centering
    \includegraphics[width=\linewidth]{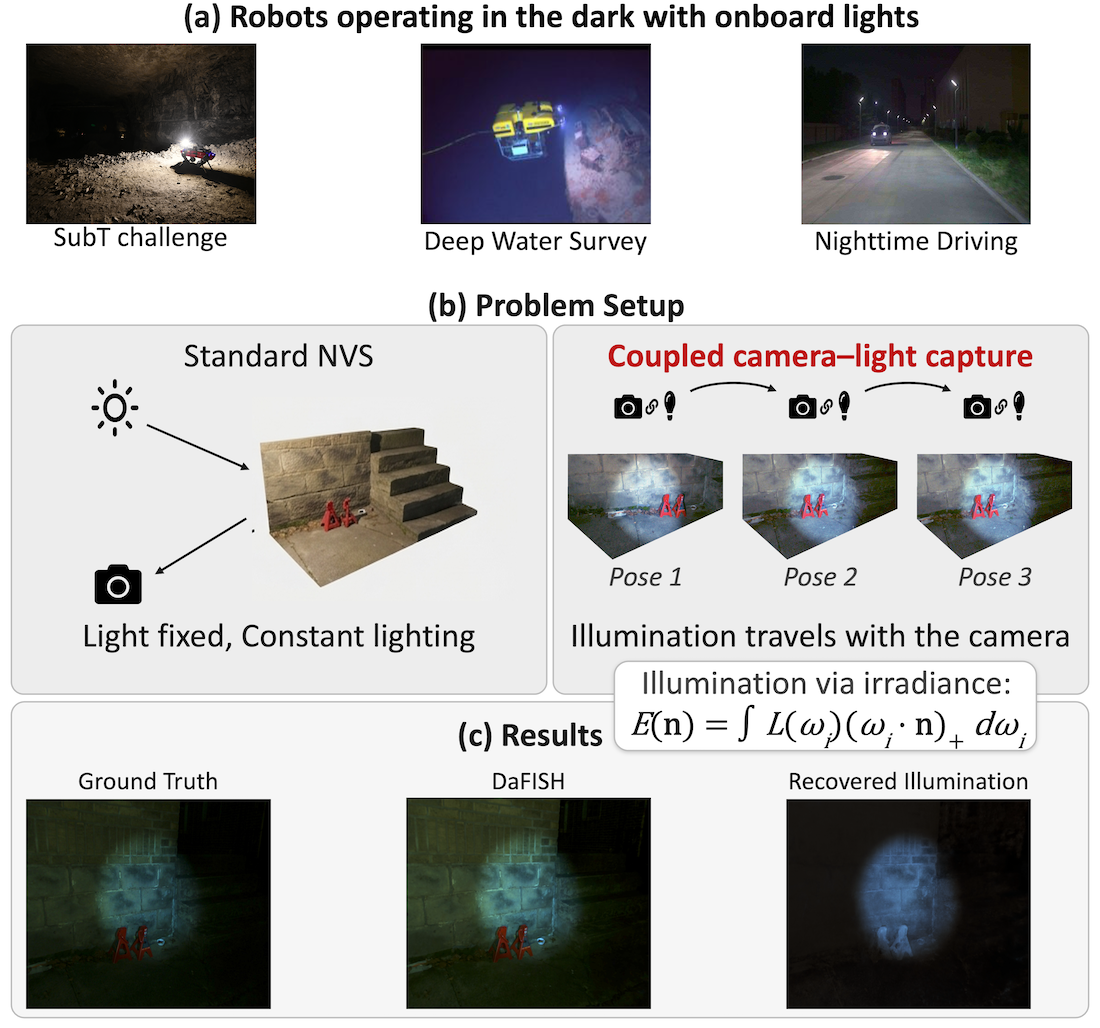}
    \caption{ \textbf{(a)} Robots that operate in dark environments, such as subterranean exploration (SubT challenge) \cite{tranzatto2022cerberus}, deep-water surveys by the remotely operated vehicle Hercules \cite{TitanicOpticsUndersea}, and nighttime driving \cite{DarkDriving} rely on onboard lights that move with the platform.
        \textbf{(b)} In standard novel view synthesis (NVS, left), hundreds of images are captured while the camera moves around the scene under fixed illumination field $L$, and a radiance field is reconstructed with methods such as NeRF, 3DGS, or their derivatives. However, in co-located capture (right), the camera and a light source are rigidly attached, so the surface illumination changes with every pose, as the light source is moving as well. 
        We model this moving illumination through the irradiance $E(\mathbf{n})$, so the irradiance for a given normal is computed by analytically rotating the coefficients with a Wigner-D matrix of the known camera motion, where the residual spatial variation from the near-field source is absorbed by a learned per-sample MLP (see \ref{sec:method} for more details).
        \textbf{(c)} DaFISH can reconstruct the scenes while attributing frame-varying brightness to the illumination branch. The brightness of ground truth and rendered images is increased by 40\% for visualization.%
    }
    \label{fig:1}
\end{figure}

\section{Introduction}
\label{sec:intro}
Neural radiance fields (NeRF)~\cite{nerf, mipnerf360} and Gaussian Splatting (GS)~\cite{tdgs,tdgs2d} have transformed novel view synthesis (NVS) and 3D scene reconstruction. However, both assume a static scene under static illumination, where the radiance emitted from a surface point is modeled as a fixed function of position and viewing direction. The incident illumination $L$ is treated as a fixed function of world coordinates, independent of camera pose or acquisition time. 

Under active illumination, the image-capture scenario violates this assumption. Whenever the light is attached to the camera, or the object moves relative to a stationary camera and light, the illumination is fixed in the camera frame, so in the scene's frame the light moves with every pose. In direct lighting capture, the light source is rigidly attached to the camera for imaging in dark, underwater, or enclosed spaces~\cite{tranzatto2022cerberus,darkgs,sunfaded,relillum}. We call this the \emph{co-located} (or co-moving~\cite{sunfaded,relillum}) lighting configuration.

Standard neural rendering lacks a mechanism to distinguish variation in local appearance due to change in illumination. This becomes a problem under a co-located light source, in which bright and dark regions occur together, as shown in Fig.~\ref{fig:1}. In such cases, as the camera repositions, the irradiance it produces on the scene shifts with every camera motion, while its angular emission profile is constant in the camera frame. As a result, each image exhibits strong, spatially localized appearance variations due to the light's pose relative to the scene surface, and the same 3D point may display entirely different colors in different views due to illumination changes. This breaks the multi-view photometric consistency that standard neural rendering relies on, leading to reconstruction artifacts.

Recent work has proposed solutions to this NVS problem. Many of these approaches construct explicit lighting or model constraints~\cite{darkgs,lightneus,prendo,reimnerf}. Other methods, like~\cite{nighttime3dgs}, make strong physical assumptions about camera or light behavior. Luminance-GS~\cite{luminancegs} uses global per-view tone curves to handle challenging lighting conditions but cannot model spatially varying illumination across an image plane. Gaussian in the Dark~\cite{gaussindark} resolves cross-view brightness inconsistency in dark captures but attributes it to the camera imaging characteristics.

\begin{figure*}[ht]
\centering
\includegraphics[width=0.8\linewidth]{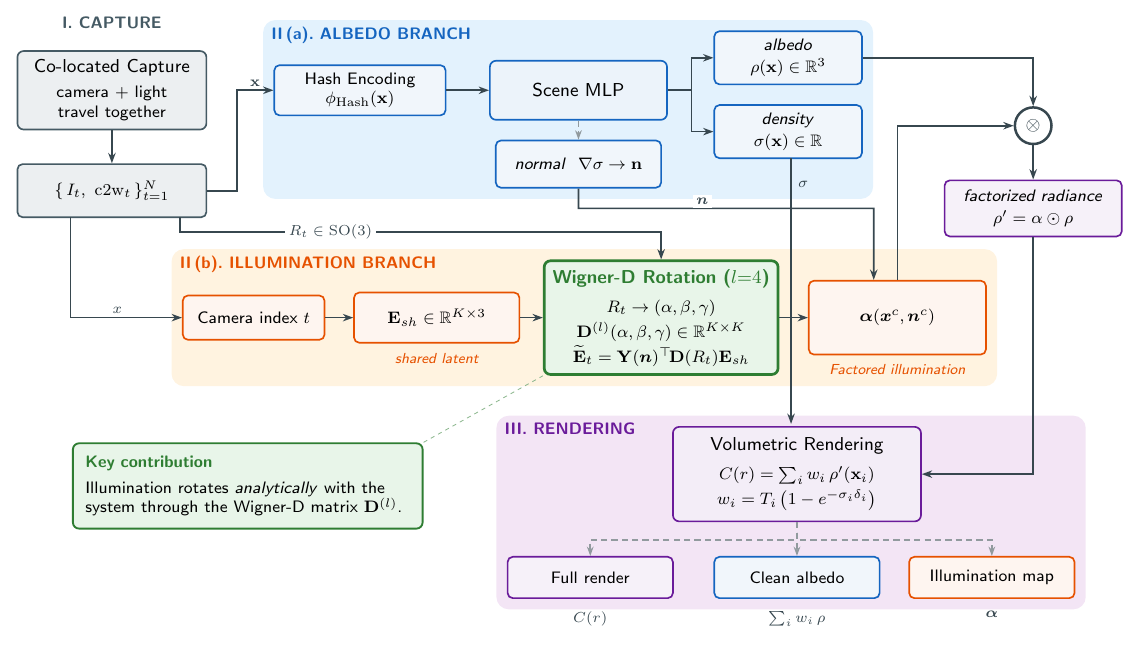}
\caption{Overview of the proposed method. Given a co-located capture that produces $N$ images $I$ and the extrinsic camera parameter matrix $c2w$ (I. Capture), the method factors appearance into albedo and illumination before volumetric rendering. The albedo branch (IIa) samples a global coordinate position $\mathbf{x}$ through a scene MLP to predict per-sample albedo $\mathbf{\rho}(\mathbf{x})$, density $\sigma(\mathbf{x})$, and a global normal $\mathbf{n}$ from the density gradient. The illumination branch (IIb) starts from a single trainable $l{=}4$ SH code $\mathbf{E}_{sh} \in \mathbb{R}^{25\times 3}$ shared across all frames, and rotates it analytically into each frame via the Wigner-D matrix $\mathbf{D}^{(l)}$ derived from the known rotation $R_t \in \mathrm{SO}(3)$, so that the rotated coefficients become $\widetilde{\mathbf{E}}_t$. Albedo $\rho(\bm{x})$ and illumination factor $\bm{\alpha}(\bm{x}^c,\bm{n}^c)$ are multiplied into the factorized radiance $\bm{\rho}' = \bm{\alpha}\odot\bm{\rho}$ and composited through volumetric rendering, producing a full render $C(\bm{r})$, a clean albedo render $\sum_i w_i\,\bm{\rho}$ (i.e., a render without illumination effects), and an illumination map $\alpha$.
}
\label{fig:overview}
\end{figure*}

In contrast to these approaches, 
we propose a formulation (DaFISH) that exploits the known rigid coupling between camera and light. As illustrated in Fig.~\ref{fig:overview}, we decompose the rendered color into an intrinsic albedo and an illumination field with irradiance, and we parameterize the irradiance in low-order spherical harmonics (SH), an orthonormal basis for functions on the sphere. The key insight, established by~\cite{ramaho_rel,ramaho_env}, is that the irradiance of a Lambertian surface is a smooth, low-order function of the normal and is well approximated by a small number of coefficients. Also, rotating a function on the sphere acts linearly and band-wise on its SH coefficients via the Wigner-D matrices~\cite{rose1995elementary,sloan2008stupid}. So, the irradiance for a given normal is obtained by an analytic Wigner-D rotation of the coefficients rather than by re-integrating over the hemisphere. This is general, only assuming a co-located lighting function, instead of a specific lighting geometry and target scene, and further does not require a well-lit reference frame. Like~\cite{relillum,hong2026spotlessgs}, it requires no calibration target and no separate calibration capture.

Our main contributions can be summarized as follows: (1) We formulate an NVS rendering method for the co-located light setting as a factorization of color into albedo and a illumination field with directional SH irradiance carried across frames by an analytic Wigner-D rotation of the known camera motion. (2) We present experiments on synthetic and real co-located captures that demonstrate better novel view reconstruction quality. (3) The formulation requires no calibration target and no separate calibration capture, making it applicable to captures already collected in the field.

\section{Related Work}
\label{sec:related}

\subsection{View-Dependent Appearance} 
Most NVS methods assume world-fixed illumination. NeRF~\cite{nerf} and Mip-NeRF~360~\cite{mipnerf360} model view-dependent color directly, and 3D/2D Gaussian Splatting~\cite{tdgs,tdgs2d} do so with explicit primitives. Ref-NeRF~\cite{refnerf} structures the view-dependent term around the reflection direction but still bakes lighting as a scene property. NeRF-W~\cite{martin2021nerf} handles illumination with a per-image appearance embedding,  and GS-W~\cite{gsw} refines this with per-Gaussian dynamic features sampled from an encoding of each reference image. Both infer each frame's lighting independently and ignore any known relationship between frames. In our setting, this relationship is known, and we exploit it analytically.

\subsection{Illumination Modeling} 
Illumination modeling has been extended to the NVS domain, often for low-light or challenging illumination scenarios. RawNeRF~\cite{mildenhall2022rawnerf} trains directly on linear raw measurements, recovering noisy and near-dark scenes. Gaussian in the Dark~\cite{gaussindark} handles brightness inconsistency in the dark via a learned camera response module. Luminance-GS~\cite{luminancegs} applies view-adaptive color-curve adjustments, and RobustSplat++~\cite{robustsplatpp} decouples illumination from geometry and dynamics in the in-the-wild setting. In both, each view's lighting is fitted independently of camera pose. We derive the \emph{frame-to-frame} lighting transform within a single capture, where the known rigid camera extrinsics make the transform \emph{analytic} rather than something to be solved for.

\subsection{Co-located Lighting} 
Several methods target capture where the light is co-located with the camera: dark-scene robot exploration~\cite{darkgs,sunfaded}, endoscopy~\cite{lightneus,prendo,reimnerf}, and underwater imaging~\cite{relillum}. In endoscopy, LightNeuS~\cite{lightneus} builds inverse-square decline into the renderer, PR-ENDO~\cite{prendo} utilizes a physically based model, and ~\cite{reimnerf} conditions the radiance field on light position. DarkGS~\cite{darkgs} additionally requires a calibration profile of the light. SpotlessGS~\cite{hong2026spotlessgs} removes the calibration step by jointly optimizing the light parameters, and uses SH to model an ambient residual. SunFaded~\cite{sunfaded} predicts per-tile lighting maps in image space from depth and light direction. We instead assume only co-located coupling with no calibration capture. Relative Illumination Fields (relative-illumination)~\cite{relillum} and SunFaded~\cite{sunfaded} factorize pixel color into albedo times an illumination factor. We adopt the same factorization but structure the illumination term such that its directional component is an SH irradiance field transported analytically via the known camera movement, rather than being learned in full.

\subsection{Irradiance and SH Rotation} 
The closed-form radiance/irradiance relationship for Lambertian surfaces can be expressed using low-order SH expansions~\cite {ramaho_rel,ramaho_env}. Wigner-D matrices are the standard tool for rotating SH signals in physics and computer graphics~\cite{rose1995elementary,sloan2008stupid}. Wang and Ramamoorthi derive analytic SH coefficients for polygonal area lights via zonal-harmonic rotation~\cite{wang2018analytic}. In neural rendering, NeRF-OSR~\cite{nerfosr} embeds an SH lighting model in a NeRF, but fits it as a world-frame environment independently for each image. In contrast, we apply the \emph{analytic} Wigner-D transport of the \emph{known} capture motion to the irradiance coefficients, yielding a nearly zero-overhead, per-frame lighting transform inside a neural radiance field.  

\section{Method}
\label{sec:method}

\subsection{Preliminaries}
\subsubsection{NeRF Rendering}
We build upon the volumetric rendering framework of NeRF~\cite{nerf}.
A scene is represented by two functions: a volume density $\sigma(\bm{x})$ of 3D position $\bm{x}$, and a directional emitted color $\bm{c}(\bm{x},\bm{d})$ of position and viewing direction $\bm{d}$. A camera ray $\bm{r}(\tau)=\bm{o}+\tau\bm{d}$ is discretized into $N$ samples $\bm{x}_i=\bm{r}(\tau_i)$, and the pixel color is obtained by alpha compositing,
\begin{equation}
\hat{C}(\bm{r})=\sum_{i=1}^{N}T_i\big(1-e^{-\sigma_i\delta_i}\big)\bm{c}_i,
~~~
T_i=\exp\Big(-\sum_{j<i}\sigma_j\delta_j\Big),
\end{equation}
where $\delta_i=\tau_{i+1}-\tau_i$ and $T_i$ is the accumulated transmittance from the camera to sample $i$.

\subsubsection{Static illumination}
For a surface point $\bm{x}$ with normal $\bm{n}$ (evaluated at the ray's expected termination and oriented along the negative density gradient), let $L(\bm{x},\bm{\omega_i})$ denote the radiance incident from direction $\bm{\omega}\in S^2=\{\bm{\omega}\in\mathbb{R}^3:\|\bm{\omega}\|=1\}$, and the bidirectional reflectance distribution function (BRDF) as $f_r(\bm{\omega_i, \omega_o})$. The outgoing radiance toward $\bm{\omega}_o$ is given by the reflection equation~\cite{kajiya1986},
\begin{equation}
B(\bm{x},\bm{n}, \bm{\omega_o})=\int_{S^2}L(\bm{x},\bm{\omega_i})f_r(\bm{\omega_i, \omega_o})\max(0,\bm{\omega}_i\cdot\bm{n})
\mathrm{d}\bm{\omega}_i,
\label{eq:irradiance_def}
\end{equation}
For a Lambertian surface with albedo $\rho(\bm{x})$ the BRDF becomes
$f_r=\rho(\bm{x})/\pi$, which can be taken outside the integral, so that
\begin{equation}
B(\bm{x},\bm{n})=\frac{\rho(\bm{x})}{\pi} E(\bm{x},\bm{n}) = \frac{\rho(\bm{x})}{\pi}\int_{S^2}L(\bm{x},\bm{\omega})
\max(0,\bm{\omega}\cdot\bm{n})\mathrm{d}\bm{\omega},
\label{eq:lambertian}
\end{equation}
which depends on the surface point and its normal, and $E$ is the irradiance. Because $L$ is frame-invariant, standard NeRF baked the illumination and albedo into a single appearance field $\bm{c}$, and therefore does not separate them. 

\subsection{Moving illumination} 
In the moving-light setting, the camera carries the illumination source. Let the camera pose at frame $t$ be $(R_t,\bm{p}_t)$, mapping canonical (camera) coordinates to world coordinates as $\bm{x}=R_t\bm{x}^c+\bm{p}_t$. The light source is fixed in the canonical frame, with free-space emitted radiance $L_c(\bm{x}^c,\bm{\omega}^c)$, and the world-frame incident radiance therefore depends on the camera pose,
\begin{equation}
L_t(\bm{x},\bm{\omega})=L_c\big(R_t^\top(\bm{x}-\bm{p}_t),R_t^\top\bm{\omega}\big).
\label{eq:Lt}
\end{equation}
Substituting $\bm{\omega}=R_t\bm{\omega}^c$ and $\bm{n}=R_t\bm{n}^c$ into the irradiance integral of Eq.~\eqref{eq:lambertian}, and using that rotations preserve inner products, $(R_t\bm{\omega}^c)\cdot(R_t\bm{n}^c)=\bm{\omega}^c\cdot\bm{n}^c$, we obtain
\begin{equation}
\label{eq:canonical_invariance}
E_t(\bm{x},\bm{n})
=\int_{S^2}L_c(\bm{x}^c,\bm{\omega}^c)\max(0, \bm{\omega}^c\cdot\bm{n}^c)\mathrm{d}\bm{\omega}^c
=E_c(\bm{x}^c,\bm{n}^c),
\end{equation}
with $\bm{x}^c=R_t^\top(\bm{x}-\bm{p}_t)$ and $\bm{n}^c=R_t^\top\bm{n}$.
The world position $\bm{x}$ and normal $\bm{n}$ of a surface point are fixed, but their canonical representations $(\bm{x}^c,\bm{n}^c)$ travel with the camera, so the irradiance at a fixed surface point varies across frames. Accordingly, in the world frame, the illumination is a family $\{L_t\}_{t=1}^T$ of pose-dependent functions, in contrast to the camera frame case, where it is the single invariant $L_c$. The per-frame radiance replacing NeRF's appearance field is
\begin{equation}
\bm{c}_t(\bm{x},\bm{d})\equiv B_t(\bm{x},\bm{n})
=\frac{\rho(\bm{x})}{\pi}E_c\big(\bm{x}^c,\bm{n}^c\big).
\label{eq:ct}
\end{equation}

\subsection{Illumination Field and Its SO(3) Transport}
\label{subsp:illu}

Based on the observation that the illumination is a pose dependent function, rather than relearning the lighting at every pose, we learn the canonical field $E_c$ and rotate it into each frame using the known camera motion. Because the angular (directional) dependence of $E_c$ transforms under an exact $SO(3)$ rotation, we can transport it analytically using Wigner-D matrices to generate a single per-frame linear map on the illumination coefficients, rather than a per-ray resampling of the light field.
We learn $E_c$ rather than $L_c$, where $E_c$ is carried by a \emph{single shared} set of SH coefficients that lives in the canonical frame, and the per-frame quantity is obtained by \emph{transporting the coefficients}, and the direction of the transport follows from the defining equivariance of the basis.

For DaFISH, the rendered directional emitted color $\bm{c}$ factorizes into an albedo $\bm{\rho}(\bm{x})$ emitted by a scene MLP and an illumination field $\bm{\alpha}$. The proposed illumination field consists of three components: an SH basis that compactly represents the irradiance, a single shared latent code that stores the canonical lighting profile once for all frames, and an analytic rotation that transports the coefficients across frames.

\subsection{SH Basis for Irradiance} 
The irradiance $E$ inherits the same SH basis evaluated at the normal, with coefficients scaled by the SH expansion of the clamped-cosine kernel $\max(0,\cos\theta)$, following~\cite{ramaho_rel} is 
\begin{equation}
E(\bm{n})=\sum_{l=0}^{l_{max}}\sum_{m=-l}^l\hat{A}_lL_{l,m}Y_{l,m}(\bm{n})
\label{eq:ramaho}
\end{equation}
where $\hat{A}_l$ are the analytic Zonal Harmonic coefficients, depending only on degree $l$ (not order $m$). The first few values are~\cite{ramaho_env}: $
\hat{A}_0=\pi,~~ \hat{A}_1=\tfrac{2\pi}{3},~~ \hat{A}_2=\tfrac{\pi}{4},
~~ \hat{A}_3=0,~~ \hat{A}_l=0 \ \text{for odd $l \geq 3$}.$
Based on the analysis from~\cite{ramaho_env} regarding the SH irradiance relationship, we use order-4 SH (25 coefficients, $l_{max}=4$) in practice to retain higher-frequency lighting detail without the computational overhead. 

\subsection{Vectorized SH Irradiance}
The scene is static in the world frame, while the illumination travels with the camera and is therefore static in the canonical (camera) frame. We represent the lighting profile by a single trainable SH coefficient vector $\mathbf{E}_{sh}=[\bm{e}_1^\top\dots\bm{e}_K^\top]^\top\in\mathbb{R}^{K \times 3}$,
$K=(l_{\max}+1)^2$, where $\bm{e}_k\in\mathbb{R}^3$ stacks the three color channels of the $k$-th irradiance coefficient $\hat{A}_lL_{l,m}$ of Eq.~\eqref{eq:ramaho}. Let $\mathbf{Y}(\bm{n}^c)\in\mathbb{R}^K$ be the real
SH basis evaluated at the canonical normal $\bm{n}^c$ (in Cartesian form).
The canonical irradiance in Eq.~\eqref{eq:canonical_invariance} is obtained in every color channels as
\begin{equation}
\mathbf{E}_c(\bm{n}^c)=\mathbf{Y}(\bm{n}^c)^\top\mathbf{E}_{sh}.
\label{eq:latent}
\end{equation}
Sharing one coefficient vector across space constrains only the angular dependence of the irradiance, which is the distant-light case. The spatially varying part of the illumination field is added by the per-sample MLP head of Sec.~\ref{subsp:factor}.

\subsection{Analytic Rotation of the Illumination Field} What changes across frames is how the fixed world normal $\bm{n}$ is seen in the camera frame, $\bm{n}^c=R_t^{\top}\bm{n}$. Rather than evaluating the canonical field at rotated normals, we act on the coefficients. The SH basis is an irreducible representation of $SO(3)$, so rotating the irradiance field by the camera/light motion $R_t$ transforms the coefficients by the unitary Wigner-D matrix $D^{(l)}(R_t)$, block-diagonal across degrees:
\begin{equation}
\begin{aligned}
\widetilde{E}_t
&=\sum_{l=0}^{l_{max}}\sum_{m=-l}^{l}E_{l,m}
Y_{l,m}\big(R_t^{\top}\bm{n}\big)\\
&=\sum_{l=0}^{l_{max}}\sum_{m=-l}^{l}
\Big(\sum_{m'=-l}^{l}D^{(l)}_{m,m'}(R_t)E_{l,m'}\Big)
Y_{l,m}\big(\bm{n}\big),
\end{aligned}
\label{eq:wigner}
\end{equation}
where $E_{l,m}$ are the entries of the shared canonical code $\mathbf{E}_{sh}$ for one color channel and $\widetilde{E}_t(\bm{n})$ is the irradiance seen at
frame $t$. Rotating the field is thus done with no re-integration over the hemisphere and no dependence on the sampling density. 

We obtain $D^{(l)}(R_t)$ from the ZYZ decomposition
$R_t=R_z(\alpha_t)R_y(\beta_t)R_z(\gamma_t)$, with $R_{ij}$ the entry in row $i$,
column $j$ of a rotation matrix,
$\beta_t=\operatorname{atan2}\big(\sqrt{R_{31}^2+R_{32}^2},R_{33}\big),
\gamma_t=\operatorname{atan2}\big(R_{32},-R_{31}\big),
\alpha_t=\operatorname{atan2}\big(R_{23},R_{13}\big).$ $D^{(l)}_{mm'}(R_t)=e^{-im\alpha_t}d^{(l)}_{mm'}(\beta_t)e^{-im'\gamma_t},$ where $d^{(l)}(\beta_t)$ is the real-valued Wigner small-$d$ matrix~\cite{rose1995elementary}. Because $D^{(l)}$ is a group homomorphism, Then, $D^{(l)}(R_1)D^{(l)}(R_2)=D^{(l)}(R_1R_2)$, and we can compose rotations in $SO(3)$ and evaluate $D^{(l)}$ once from $R_t$. We handle the gimbal lock at $\sin\beta_t=0$ by setting $\alpha_t=0$ and $\gamma_t$ from the upper-left $2\times2$ block of $R_t$. The relationship between real and complex SH is given by a fixed unitary change of basis, so the real coefficients are rotated by converting to the complex basis, applying $D^{(l)}$, and converting back~\cite{sloan2008stupid}.

We can then write down $\widetilde{E}_t$ as a matrix multiplication. For a sample with world normal $\bm{n}$, the per-frame irradiance then becomes
\begin{equation}
\widetilde{\mathbf{E}}_t=\mathbf{Y}(\bm{n})^\top
\mathbf{D}(R_t)\mathbf{E}_{sh}.
\label{eq:latent_share}
\end{equation}
where $\mathbf{Y}(\bm{n})\in\mathbb{R}^{K}$ stacks the SH basis values, $\mathbf{E}_{sh}$ stacks the coefficients of all three color channel. The coefficient vector $\mathbf{E}_{sh}$ is a \emph{single shared} parameter vector used at every frame, and the frame index $t$ enters only through the rotation $\mathbf{D}(R_t)$, which is shared across all rays and samples.

\begin{table*}[h]
\centering\small
\caption{NVS results on five co-located captures of the DarkRobotic dataset~\cite{darkgs} and FourLights ~\cite{relillum}. The relative-illumination baseline~(\cite{relillum}) failed the Outdoor2 reconstruction before reaching the maximum iteration and is reported with n/a there. Best per scene in \textbf{bold}, second-best in \emph{italics}.}
\label{tab:main}
\resizebox{\linewidth}{!}{\begin{tabular}{lcccccccccccccccccc}
\toprule
 & \multicolumn{3}{c}{Lab1} & \multicolumn{3}{c}{Lab2} & \multicolumn{3}{c}{Outdoor1} & \multicolumn{3}{c}{Outdoor2} & \multicolumn{3}{c}{FourLights} \\
\cmidrule(lr){2-4}\cmidrule(lr){5-7}\cmidrule(lr){8-10}\cmidrule(lr){11-13}\cmidrule(lr){14-16}
Method & PSNR$\uparrow$ & SSIM$\uparrow$ & LPIPS$\downarrow$  & PSNR$\uparrow$& SSIM$\uparrow$ & LPIPS$\downarrow$  & PSNR$\uparrow$ & SSIM$\uparrow$ & LPIPS$\downarrow$  & PSNR$\uparrow$ & SSIM$\uparrow$& LPIPS$\downarrow$  & PSNR$\uparrow$ & SSIM$\uparrow$ & LPIPS$\downarrow$  \\
\midrule
2DGS~\cite{tdgs2d} & 28.52 & 0.9363 & 0.1629 & 29.70 & 0.9212 & 0.1844 & 26.23 & 0.9268 & 0.2330 & 36.97 & 0.9352 & 0.1717 & 27.28 & 0.7777 & 0.1644 \\
3DGS~\cite{tdgs} & 30.08 & 0.9463 & 0.1017 & 29.82 & 0.8626 & 0.1697 & 25.17 & 0.8723 & 0.2117 & 37.86 & 0.9505 & \emph{0.0920} & 27.30 & 0.7637 & 0.1620 \\
GS-W~\cite{gsw} & 37.03 & \emph{0.9582} & 0.0851 & 29.14 & 0.8444 & 0.2393 & 25.69 & 0.8685 & 0.2238 & 33.47 & 0.8251 & 0.1508 & 27.84 & 0.8778 & 0.2043 \\
DarkGS~\cite{darkgs} & \emph{38.15} & 0.9508 & 0.1249 & 39.03 & 0.9587 & 0.1385 & \emph{35.25} &0.9546 & 0.1909 & \emph{41.22} & 0.9506 & 0.1708 & \multicolumn{3}{c}{n/a} \\
SpotlessGS~\cite{hong2026spotlessgs}  & 35.01 & 0.9573 & 0.1107 & 36.80 &\emph{0.9642} & 0.1236 & 30.44 & \emph{0.9584} & 0.1683 & 40.93&  \emph{0.9595} & 0.1578 &25.51 & 0.8182&0.1720\\
relative-illumination~\cite{relillum} & 37.78 & 0.9419 & \emph{0.0799} & \emph{41.90} & 0.9622 & \emph{0.0533} & 33.67 & 0.9321 & \emph{0.1563} & \multicolumn{3}{c}{n/a} & \textbf{29.57} & \emph{0.8948} & \emph{0.1388} \\
DaFISH & \textbf{39.57} & \textbf{0.9607} & \textbf{0.0569} & \textbf{42.45} & \textbf{0.9741} & \textbf{0.0418} & \textbf{38.71} & \textbf{0.9687} & \textbf{0.0667} & \textbf{45.34} & \textbf{0.9756} & \textbf{0.0375} & \emph{28.46} & \textbf{0.9276} & \textbf{0.0880} \\
\bottomrule
\end{tabular}}
\end{table*}
\subsection{Albedo$\times$Irradiance Factorization and Rendering}
\label{subsp:factor}
We represent the scene by the scene MLP that outputs density $\sigma(\bm{x})$ and albedo $\bm{\rho}(\bm{x})$ at each ray sample in the world coordinate frame. For frame $t$ and ray $\bm{r}(\tau)=\bm{o}_t+\tau\bm{d}_t$, the rendered color is the volume integral of albedo modulated by the per-sample illumination factor  $\bm{\alpha}(\bm{x}^c,\bm{n}^c)$:
\begin{equation}
\hat{C}(\bm{r})=\sum_{i=1}^{N}T_i\big(1-e^{-\sigma_i\delta_i}\big)
\bm{\rho}(\bm{x}_i)\odot\bm{\alpha}(\bm{x}^c_i,\bm{n}^c_i),
T_i=e^{-\sum_{j<i}\sigma_j\delta_j}.
\label{eq:render}
\end{equation}
The product $\bm{\rho}\odot\bm{\alpha}$ separates intrinsic appearance from illumination. The illumination factor combines the frame-$t$ irradiance $\widetilde{E}_t$ of Eq.~\eqref{eq:latent_share} with a per-sample MLP head $\mathcal{F}^{\ell}_\Theta$ that takes a hash-grid encoding of the canonical position and an SH feature encoding of the canonical normal, for extending the illumination field to a spatially-varying field. The two factors are multiplied together inside a logistic sigmoid,
\begin{equation}
\begin{aligned}
\bm{\alpha}(\bm{x}^c,\bm{n}^c)=\operatorname{sig}\Big(&
\SHshift\widetilde{E}_t\mathcal{F}^{\ell}_{\Theta}\big(\phi_{\mathrm{Hash}}(\bm{x}^c),\\
&\phi_{\mathrm{SH}}(\bm{n}^c)\big)-\ashift\Big),
\end{aligned}
\label{eq:alpha_mlp}
\end{equation}
where $\operatorname{sig}(\cdot)$ is the logistic sigmoid. $\mathcal{F}^{\ell}_{\Theta}$ is a 2-layer MLP head, $\phi_{\mathrm{Hash}}$ the multi-resolution hash-grid feature at the canonical position, and $\phi_{\mathrm{SH}}$ an SH positional encoding of the normal, distinct from
the irradiance basis $\mathbf{Y}$. We utilize \cite{muller2022instantngp} for the hash-grid encoding. The gain scalar $\SHshift$ and the offset $\ashift$ act on the pre-activation, so that at initialization the illumination branch starts from a controlled value that matches the darkness of the capture.

\subsection{Training Loss}
\label{sec:training}

The objective combines a rendering term with frequency-aware SH regularization,
\begin{equation}
\mathcal L= \sum_{r \in R} \left\| \frac{\hat{C}(r(t)) - C(r(t))}{\operatorname{sg}(\hat{C}(r(t))) + \epsilon} \right\|_2^2 +\lambda_{\mathrm{sh}}\mathcal L_{\mathrm{sh\_reg}},
\end{equation}
and the $\ell_2$ term is from~\cite{relillum}, as ground truth $ C(r(t))$ and prediction $\hat{C}(r(t))$ are reweighted coordinate-wise with $sg(\cdot)$ as the stop-gradient operator and $\epsilon=10^{-3}$. The SH regularization penalizes high-frequency coefficients of $\widetilde{\mathbf{E}}_t$ proportional to their degree~$l$:
\begin{equation}
\mathcal L_{\mathrm{sh\_reg}}=\frac{1}{(l_{max}+1)^2}\sum_{l=0}^{l_{max}}\sum_{m=-l}^{l} l \|\widetilde{\mathbf{E}}_t^{l,m}\|_2^2,
\end{equation}
which discourages the model from relying on high-order SH modes. 

\begin{figure*}[h]
\centering
\includegraphics[scale=0.75]{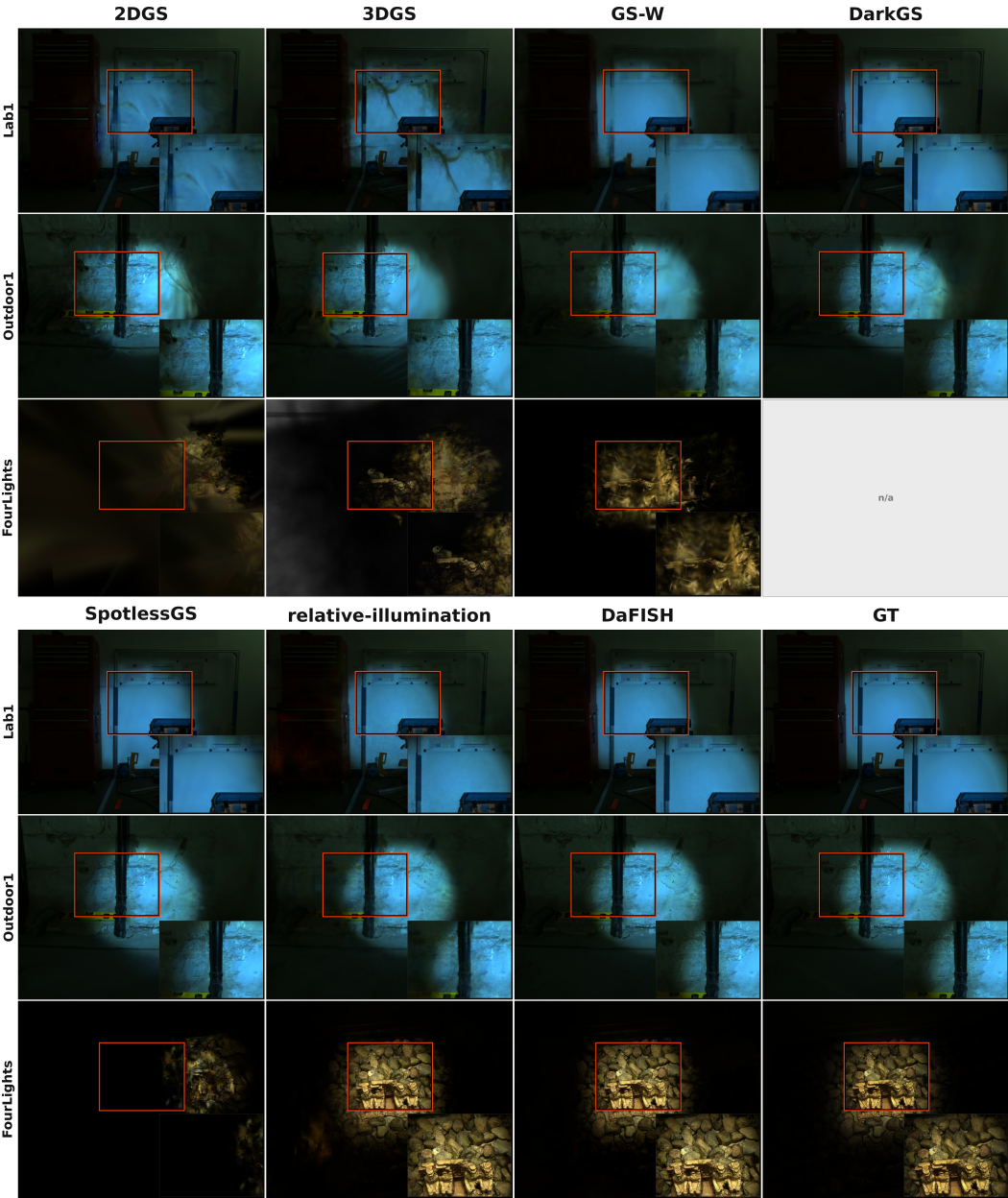}
\caption{Qualitative comparison across three co-located captures (rows: Lab1,  Outdoor1, and FourLights) and all methods plus ground truth (GT) are split over two stacked panels of four columns (top 4: 2DGS~\cite{tdgs2d}, 3DGS~\cite{tdgs}, GS-W~\cite{gsw}, DarkGS~\cite{darkgs}; bottom 4: SpotlessGS~\cite{hong2026spotlessgs}, relative-illumination~\cite{relillum}, DaFISH, and GT). The red bounding box highlights the reconstructed detail. Our factorization keeps albedo stable while the directional SH irradiance travels with the camera.}
\label{fig:qual}
\end{figure*}

\section{Experiments}
\label{sec:experiments}
\subsection{Datasets} We evaluate on the public DarkRobotic dataset~\cite{darkgs}, captured with a FLIR Firefly S camera and an active light source co-located rigidly on a legged robot, and the synthetic data from relative-illumination~\cite {relillum}. Two datasets comprise a total of five co-located sequences. For each sequence, we hold out every 8th frame, serving as the validation set on which all metrics are reported, and the remaining frames are used for training.

\subsection{Baselines} We compare our proposed method against 2D Gaussian Splatting (2DGS)~\cite{tdgs2d}, 3D Gaussian Splatting (3DGS)~\cite{tdgs}, GS-W~\cite{gsw}, DarkGS~\cite{darkgs}, SpotlessGS~\cite{hong2026spotlessgs}, and a NeRF-based model relative-illumination~\cite{relillum}. SunFaded~\cite{sunfaded} is not open-source at the time of this report.

\subsection{Metrics} We report three standard image-quality metrics on the held-out validation set. \emph{PSNR} (dB) measures pixel-wise fidelity via the mean squared error between the render and the reference. \emph{SSIM}~\cite{wang2004ssim} ($[0,1]$) captures perceived structural similarity, where 1 is the perfect match.  \emph{LPIPS}~\cite{zhang2018lpips} is a perceptual distance computed between deep features of the two images. All three are averaged over the validation frames of each sequence.  

\subsection{Training Setup} 
We implement our method in PyTorch on top of the relative-illumination codebase \cite{relillum} and train all sequences for 30{,}000 iterations on a single NVIDIA RTX 5090 GPU, with each scene taking roughly 16 minutes. All parameters, including the scene field, the illumination field, and the SH coefficients, were optimized jointly with Adam~\cite{adam}. The learning rate is set to $10^{-2}$ for both the scene and illumination fields and the SH latent vector, except $10^{-3}$ for the Outdoor2 scene. We sample 4{,}096 rays per batch. We set $\ashift$ as a scale related to the darkness. We use $\ashift=2.5$ for outdoor scenes (Outdoor1, Outdoor2, and FourLights), and $0.0$ for the rest (Lab1, Lab2). We also set $\lambda_{\mathrm{sh}} = 0.45$ and $\SHshift=0.65$ for all reported experiments. For $\phi_{\mathrm{Hash}}$, we use size=17 for the hash map.

\section{Results}
\label{sec:results}

\subsection{Quantitative Results}\footnote{DarkGS~\cite{darkgs} is not evaluated on the FourLights scene, as its light-simulator components require calibration parameters of the light, which \cite{relillum} does not provide (n/a). Because Outdoor2 is the one scene requiring our reduced learning rate of $10^{-3}$, we also tried~\cite{relillum} with this setting. It still failed before the maximum iteration.} 
Table~\ref{tab:main} summarizes the comparison on four co-located captures (Lab1, Lab2, Outdoor1, and Outdoor2) from the DarkRobotic dataset~\cite{darkgs}, plus FourLights from relative-illumination~\cite{relillum}. Our method DaFISH achieves the best SSIM and LPIPS on all five captures and the best PSNR on all four DarkRobotic scenes, outperforming the strongest reported splatting baseline, DarkGS~\cite{darkgs} in terms of PSNR, SpotlessGS~\cite{hong2026spotlessgs} for SSIM and LPIPS, and the closest NeRF-based baseline, relative-illumination~\cite{relillum}. On FourLights, the splatting baselines degrade most under the moving light, whereas our factorization stays consistent across all five scenes (PSNR $38.71$ to $45.34$~dB on DarkRobotic and $28.46$~dB on FourLights). The relative-illumination baseline~\cite{relillum} leads in PSNR ($29.57$~dB), while we retain clearly better SSIM ($0.9276$ vs.\ $0.8948$) and LPIPS ($0.0880$ vs.\ $0.1388$), showing our method adapts well to this setting and maintains competitive performance for NVS.

\subsection{Qualitative Results} 
\begin{figure}[t]
\centering
\includegraphics[width=0.88\linewidth]{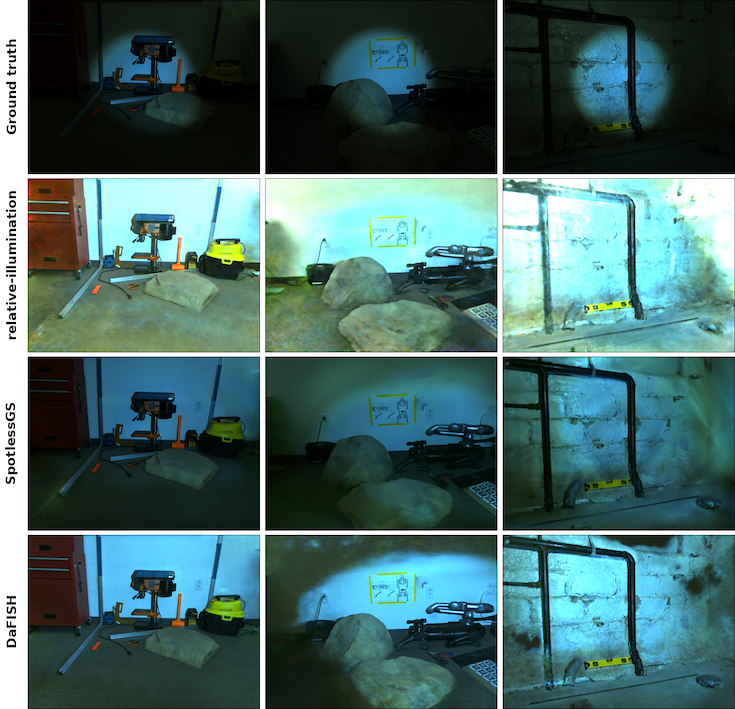}
\caption{Clean scene recovery on held-out frames across three DarkRobotic scenes (columns). For each scene, a held-out GT frame (top) is shown next to the clean albedo (no illumination effect render) of the baselines (SpotlessGS~\cite{hong2026spotlessgs}, relative-illumination~\cite{relillum}) and DaFISH.}
\label{fig:albedo}
\end{figure}

\begin{figure}[t]
\centering
\includegraphics[width=0.80\linewidth]{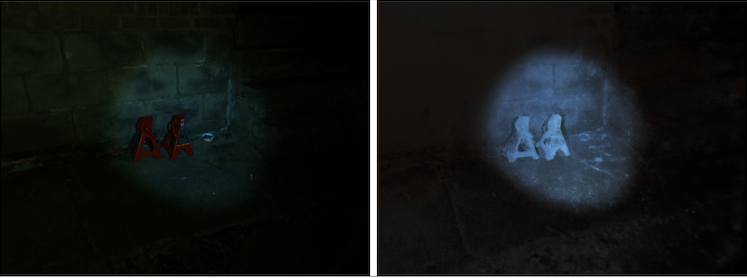}
\caption {The row shows the original capture (left) next to the learned illumination field (right) at the same camera orientation.}
\label{fig:illum}
\end{figure}

Fig.~\ref{fig:qual} extends the visual comparison to all baselines. Most baseline splatting methods, except GS-W, DarkGS, and SpotlessGS, bake frame-varying illumination into the texture, producing artifacts where the scene is exposed by moving light. The zoom insets under each method highlight the reconstructed details. DarkGS, SpotlessGS, and relative-illumination produce visually plausible results of DarkRobotic, but the results are blurry. For FourLights, only relative-illumination and DaFISH could produce results comparable to the GT. Fig.~\ref{fig:albedo} extends the visual comparison for clean albedo recovery with SpotlessGS and relative-illumination. Both relative-illumination and DaFISH reuse the recovered albedo without the illumination field. Compared to SpotlessGS and relative-illumination, DaFISH recovers finer texture detail. Even though relative-illumination and DaFISH separate appearance into albedo and illumination, relative-illumination exhibits strong over-exposure. In Fig.~\ref{fig:illum}, we show the learned illumination field according to the GT capture at the same camera orientation. The recovered maps concentrate in the directions of the co-located spotlights and follow the illumination profile, confirming the illumination branch's ability to learn the co-located lighting.

\subsection{Ablation Study}
Table~\ref{tab:ablation} isolates the contribution of each component of DaFISH on Outdoor1 and FourLights under the same hold-out protocol, changing exactly one setting at a time from the full model.

\subsubsection{Influence of the Rotated-Irradiance Term} 
DaFISH becomes the baseline relative-illumination~\cite{relillum} when dropping the $\widetilde{E}_t$ from Eq.~\eqref{eq:alpha_mlp}. For comparison with Outdoor1, PSNR falls by 5.04~dB, SSIM falls by $0.0366$, and LPIPS more than doubles ($0.1563$ vs. $0.0667$).

\subsubsection{Influence of the SH order} We sweep the SH band $l=1,2, 4$ with FourLights. As shown in Table~\ref{tab:ablation}, dropping to $l{=}1$ degrades PSNR by $2.63$~dB. Order 4 therefore covers the meaningful irradiance content: $\textbf{A}$ decays as $l^{-5/2}$ and vanishes at odd orders above one ~\cite{ramaho_rel,ramaho_env}. We set $l{=}4$ for all reported experiments. 

\section{Conclusion and Limitation}
\label{concl}
We presented DaFISH, a plugin component for NVS under co-located illumination. By factorizing per-point emitted colors into surface albedo and an illumination factor parameterized by SH, we learn the frame-dependent illumination component caused by traveling light as an explicit, pose-transported illumination term rather than a scene field. The Wigner-D rotation of the SH coefficients handles the per-view illumination change analytically through a closed-form SO(3) operation, keeping the whole pipeline stable for joint optimization of albedo, lighting, and geometry. On five co-moving captures across the DarkRobotic and synthetic datasets, DaFISH achieves an average PSNR of 38.91~dB, SSIM of 0.9613, and LPIPS of 0.0582, and yields more consistent reconstruction across views. While our method currently uses NeRF-based volume rendering, the SH irradiance factorization is representation-agnostic. Integrating it with Gaussian splatting would enable real-time rendering under co-located illumination, enabling more robust downstream robotic perception and complementing SunFaded~\cite{sunfaded}, DarkGS~\cite{darkgs}, and SpotlessGS~\cite{hong2026spotlessgs}. 

\textbf{Limitations.} The offset $\ashift$ that sets the initialization of the illumination sigmoid is hand-picked per environment, and its sensitivity remains unexplored. A learned or data-driven initialization would remove this manual choice. While the factorized albedo and the learned illumination field are theoretically separable, practical relighting under arbitrary (non-capture) illumination depends on the completeness of the irradiance sampling during capture. 

\textbf{Future work.} We will explore environment map extraction from the learned SH coefficients and evaluate it under synthetic and real relighting conditions. Also, our current formulation assumes Lambertian (diffuse) reflectance. Highly reflective surfaces would require explicit specular modeling, e.g.\ view-dependent specular terms observed in endoscopy. 

\begin{table}[t]
\centering\small
\caption{Ablation of the DaFISH components on Outdoor1 and FourLights.}
\label{tab:ablation}

\begin{tabular}{@{}lccc@{}}
\toprule
Variation & PSNR $\uparrow$ & SSIM $\uparrow$ & LPIPS $\downarrow$\\
\midrule
Full model   (Outdoor1)                           & 38.71 & 0.9687 & 0.0667 \\
relative-illumination (Outdoor1)                  & 33.67 & 0.9321 & 0.1563 \\
\midrule
SH order $l{=}1$  (FourLights)                    & 25.83 & 0.9148 & 0.0954 \\
SH order $l{=}2$  (FourLights)                    & 27.30 & 0.9197 & 0.0924 \\
SH order $l{=}4$  (FourLights)                    & 28.46 & 0.9276 & 0.0880 \\
\midrule
\end{tabular}
\end{table}


\bibliographystyle{ieeetr}
\bibliography{directional_illumination_factorization}


\end{document}